\documentclass[10pt,twocolumn,letterpaper]{article}

\usepackage{iccv}              
\usepackage{enumitem}
\usepackage{afterpage}
\usepackage{amsmath}
\usepackage{tabularx}
\usepackage{algorithm} 
\usepackage{algpseudocode} 
\usepackage{placeins}
\usepackage{stfloats}
\usepackage{needspace}

\newcolumntype{U}{>{\centering\arraybackslash}p{1.7cm}}
\newcolumntype{V}{>{\centering\arraybackslash}p{0.8cm}}
\newcolumntype{W}{>{\centering\arraybackslash}p{1.1cm}}
\newcolumntype{Y}{>{\arraybackslash}p{2.15cm}}
\newcolumntype{Z}{>{\centering\arraybackslash}p{0.55cm}}

\newcolumntype{A}{>{\centering\arraybackslash}p{0.5cm}}
\newcolumntype{B}{>{\arraybackslash}p{0.55cm}}

\usepackage{pifont}
\newcommand{\cmark}{\ding{51}}%
\newcommand{\xmark}{\ding{55}}%

\newcommand\blfootnote[1]{%
  \begingroup
  \renewcommand\thefootnote{}\footnote{#1}%
  \addtocounter{footnote}{-1}%
  \endgroup
}

\definecolor{cvprblue}{rgb}{0.21,0.49,0.74}
\usepackage[pagebackref,breaklinks,colorlinks,allcolors=cvprblue]{hyperref}

\def\paperID{9716} 
\def\confName{ICCV}
\def\confYear{2025}

\title{GeoDiffusion: A Training-Free Framework for Accurate 3D Geometric Conditioning in Image Generation}
\vspace{-24pt}
\author{Phillip Mueller$^{1,2,*}$ \quad 
Talip Uenlue$^{1,3}$ \quad
Sebastian Schmidt$^{2,3}$ \quad \\
Marcel Kollovieh$^{3}$ \quad 
Jiajie Fan$^{1,4}$ \quad
Stephan Günnemann$^{3}$ \quad
Lars Mikelsons$^{1}$\\
$^1$ University of Augsburg \quad $^2$ BMW Group \quad $^3$ Technical University of Munich \quad $^4$ Leiden University}

\begin{document}
\twocolumn[{
\maketitle
\vspace{-24pt}
\begin{center}
    \centering
    \captionsetup{type=figure}
    \includegraphics[width=0.9\linewidth]{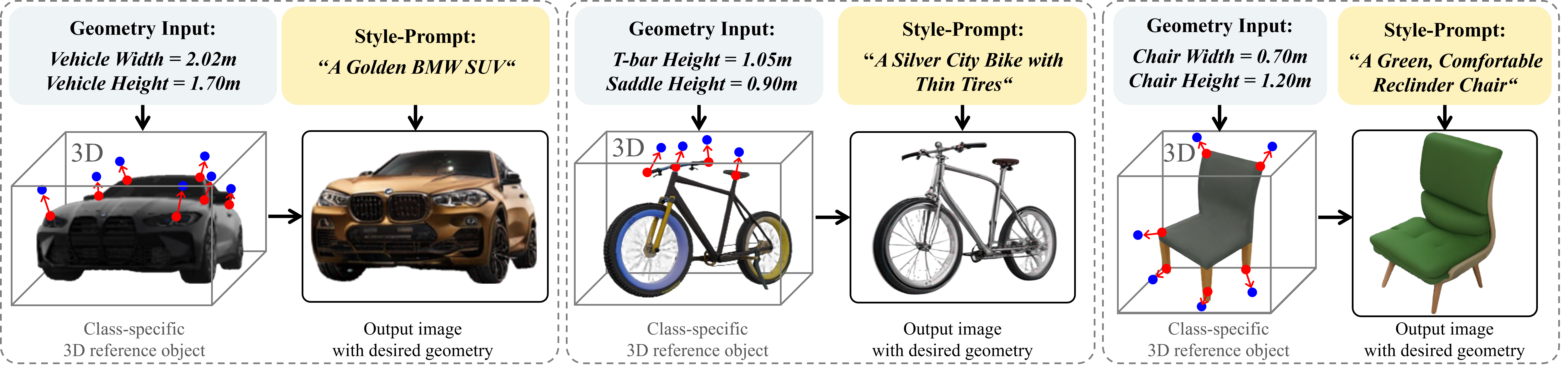}
    \vspace{-0.5em}
    \caption{Geometric control of 3D object features in image generation with \textbf{GeoDiffusion}. Given a class-specific 3D reference object (left), the target geometric characteristics are defined in 3D as parameterizable equations and projected as pairs of source and target keypoints. The geometric constraints are projected into the 2D viewpoint and together with the style-prompt, the geometry-guided image is generated.}
    \vspace{-0.5em}
    \label{fig:teaser}
\end{center}
}]
\maketitle
\begin{abstract}
Precise geometric control in image generation is essential for engineering \& product design and creative industries to control 3D object features accurately in image space. Traditional 3D editing approaches are time-consuming and demand specialized skills, while current image-based generative methods lack accuracy in geometric conditioning. To address these challenges, we propose \textbf{GeoDiffusion}, a training-free framework for accurate and efficient geometric conditioning of 3D features in image generation. GeoDiffusion employs a class-specific 3D object as a geometric prior to define keypoints and parametric correlations in 3D space. We ensure viewpoint consistency through a rendered image of a reference 3D object, followed by style transfer to meet user-defined appearance specifications. 
At the core of our framework is \textbf{GeoDrag}, improving accuracy and speed of drag-based image editing on geometry guidance tasks and general instructions on DragBench. Our results demonstrate that GeoDiffusion enables precise geometric modifications across various iterative design workflows.
\blfootnote{Contact author: Phillip Mueller (\textit{phillip.mueller@uni-a.de}).}
\end{abstract}    
\vspace{-0.5em}
\section{Introduction}
\label{sec:intro}
Visual images are essential for exchanging ideas and exploring design options in fields such as engineering design and creative industries like game design and advertising. Although real-world designs are typically created in 3D, 2D images remain more practical for concept exchange and rapid feedback. However, converting a 3D model to a 2D image inherently loses spatial information, making it challenging to adjust 3D geometric features accurately within the image. The same geometric features of an object can be vastly different in image space, depending on the viewpoint. Achieving precise geometric control over object features traditionally requires specialized software, technical expertise, and tools like Photoshop \cite{AdobePhotoshop}, Blender \cite{Blender} or AfterEffects \cite{AdobeAfterEffects}. While direct 3D editing could address these issues, it requires extensive manual labor, limiting its practicality in early design phases where ease of quick iterations and rapid generation of ideas are essential.

Generative models with precise geometric control have the potential to increase both accessibility and efficiency of design-object generation and manipulation in professional and creative contexts. However, achieving such precision remains a significant challenge, even with recent advancements in generative models for image synthesis \cite{sohldickstein2015deepunsupervisedlearningusing, ho2020denoisingdiffusionprobabilisticmodels, song2022denoisingdiffusionimplicitmodels, dhariwal2021diffusionmodelsbeatgans, rombach2022highresolutionimagesynthesislatent, podell2023sdxlimprovinglatentdiffusion} and conditional modifications \cite{zhang2023addingconditionalcontroltexttoimage, voynov2022sketchguidedtexttoimagediffusionmodels, shi2023dragdiffusionharnessingdiffusionmodels, mou2023t2iadapterlearningadaptersdig, tumanyan2022plugandplaydiffusionfeaturestextdriven,kollovieh2023predict}. 

Existing image generation approaches enable conditional control through modalities like text, reference images, or spatial guiding inputs. Text-based image generation \cite{rombach2022highresolutionimagesynthesislatent, hertz2022prompt, tumanyan2022plugandplaydiffusionfeaturestextdriven, brooks2022instructpix2pix, kawar2023imagic} is suitable for general guidance and straightforward manipulations like style changes. However, specific and technical instructions typically fail because they require a domain-specific \emph{object understanding} in the image and nViecessitate incorporating precise metric scales.
Inpainting \cite{rombach2022highresolutionimagesynthesislatent, Lugmayr_2022_CVPR}, sketch-guidance \cite{voynov2022sketchguidedtexttoimagediffusionmodels} and trainable adapters for spatial conditioning  \cite{zhang2023addingconditionalcontroltexttoimage, mou2023t2iadapterlearningadaptersdig} offer geometric control but lack precision and intuitive conditioning modalities like geometric reference points. Trainable adapters often require extensive training data and computational resources, making them difficult to customize. 

Drag-based editing provides interactive spatial control by enabling image modifications through the selection of seed and target points \cite{pan2023dragganinteractivepointbased, shi2023dragdiffusionharnessingdiffusionmodels, mou2023dragondiffusionenablingdragstylemanipulation,zhang2024gooddraggoodpracticesdrag, hou2024easydrag, nie2024blessingrandomnesssdebeats, liu2024dragnoiseinteractivepointbased}. The intuitive formulation of the modification instructions as point-pairs allows for unmatched customization, especially in integrated image generation frameworks. However, existing dragging methods exhibit inaccuracies in geometric modifications (the selected points not reaching their intended target), long inference times, and 
unintended alterations to the object's identity in the image. Crucially, they lack the ability to specify and verify \emph{precise displacements of 3D object features}. Consequently, purely 2D dragging cannot guarantee geometry-aware editing or generation on its own. 

Overall, existing generative methods lack precise control over 3D geometrical features of objects in image generation, making it impossible to enforce structural constraints with accuracy. To achieve true adherence to geometric constraints in image space, the generation must be guided by 3D-informed priors that dictate the object’s real-world dimensions and proportions. There remains a gap in integrated frameworks that provide accurate, efficient, and user-friendly geometric control without requiring extensive training data. To address this limitation, we propose \textbf{GeoDiffusion}\footnote{Code is found under: \url{https://github.com/Phillip-M97/ICCV_GeoDiff}}, a novel training-free framework that enables exact 3D-aware geometric control in image generation. We target early-phase visual concept design, a fundamental early step in product development cycles. The primary application involves rapidly
sketching images of product designs without an existing 3D product model. 

By leveraging a reference object’s actual 3D composition and the user-defined target geometry as a 3D-informed prior, GeoDiffusion translates parametric constraints directly into a guidance signal for 2D image generation. We derive the 3D-informed prior by first computing the difference between a reference object’s 3D geometric features and the user-defined target geometry. We subsequently project this difference onto the 2D image plane, resulting in viewpoint-accurate source–target keypoint pairs that serve as explicit instructions for our drag-based editing procedure \textbf{GeoDrag}. Unlike purely 2D conditioning methods, the GeoDiffusion framework ensures that the generated representation in the image precisely reflects the intended 3D structural features and constraints of the object under changing viewpoints.

GeoDiffusion divides the generative process into four sequential steps: \textbf{(1)} It begins with a single, domain-specific 3D reference object loaded into a \textit{Blender} scene. Within the 3D environment, the user sets the geometric keypoints that describe the base geometry (source points) and defines the point translation function, which establishes how modifications affect these keypoints. The target keypoints can be either manually specified or computed automatically using this function, which encodes the relationships between object features and their transformations. For example, in the case of a car, if source keypoints mark the wheel centers and the front and rear ends, the point translation function determines how these points shift when adjusting the car’s length, ensuring structural consistency in the modification. The user then selects the viewpoint to render an image of the object, and projects the source and target keypoints into image space. \textbf{(2)} The \textit{PnP Diffusion Features} method~\cite{tumanyan2022plugandplaydiffusionfeaturestextdriven} is employed to modify the object according to an input prompt. \textbf{(3)} Geometric modifications to achieve the target geometry are performed with our improved dragging approach, \textbf{GeoDrag}, which increases accuracy and efficiency compared to existing methods, especially when handling multiple reference points. \textbf{(4)} The modified image is refined using \textit{SDXL}-refiner \cite{podell2023sdxlimprovinglatentdiffusion} to enhance image fidelity. Our key contributions are:
\begin{itemize}
  \item With \textbf{GeoDiffusion}, we introduce a framework that enables \emph{accurate geometric conditioning with precision measurements in image generation} (\Cref{fig:teaser}). To our knowledge, \textbf{GeoDiffusion} is the first framework that unifies 3D‐informed geometric control with 2D diffusion‐based editing. Leveraging a single 3D object, our approach is highly adaptable for custom domains. We term it \emph{training-free} as there is only \emph{minimal LoRA fine-tuning} involved that takes less than 10 seconds per object. 
  \item We further propose \textbf{GeoDrag} to \emph{enhance accuracy and speed of dragging-based image editing} by introducing a point fixation mechanism and integrating latent optimization with a copy-and-paste strategy. While it can serve as a standalone solution for faster and more precise drag-based image editing, integrating it into GeoDiffusion enables 3D-constrained image generation.
  \item We evaluate GeoDiffusion across different object categories, introducing a new dataset for geometry-guided image generation. On this dataset, we demonstrate consistent image generation under 3D constraints. Additionally, we benchmark GeoDrag and existing dragging methods in terms of speed, accuracy, and image fidelity.
\end{itemize}


\section{Related Work}
\label{sec:Related_Work}

\noindent \textbf{Spatial Guidance in Image Generation.} ControlNet introduces adapters to guide diffusion models using control signals like sketches or depth maps \cite{zhang2023addingconditionalcontroltexttoimage}. However, it requires extensive training (200k data-pairs and 300 hours of training for an adapter). T2I-Adapter \cite{mou2023t2iadapterlearningadaptersdig} learns adapters for spatial conditioning by training separate weights and injecting them into Stable Diffusion, with the same drawbacks as ControlNet. Sketch-Guided Diffusion \cite{voynov2022sketchguidedtexttoimagediffusionmodels} allows conditioning on sketch inputs but is constrained to processing sketch-images and, therefore, cannot handle arbitrary geometric reference points specified by the user.

\noindent\textbf{Dragging-based Image Editing.} Dragging-based image editing methods \cite{pan2023dragganinteractivepointbased, shi2023dragdiffusionharnessingdiffusionmodels} enable users to select seed and target points, manipulating the image content through iterative optimization of latent features. DragDiffusion involves a three-step process: identity-preserving fine-tuning using LoRA \cite{hu2021loralowrankadaptationlarge}, latent optimization, and reference-latent control to maintain image identity. GoodDrag \cite{zhang2024gooddraggoodpracticesdrag} improves image quality by alternating between dragging and denoising steps, optimizing the latent over multiple timesteps. SDE-Drag \cite{nie2024blessingrandomnesssdebeats} proposes a simple dragging method by copying and pasting the source point in the latent image to its target position and blurring the source position.
Other recent approaches include DragonDiffusion \cite{mou2023dragondiffusionenablingdragstylemanipulation}, FreeDrag \cite{ling2023freedragfeaturedraggingreliable}, DragNoise \cite{liu2024dragnoiseinteractivepointbased} and EasyDrag \cite{hou2024easydrag}.

\noindent\textbf{3D-Based Modification Frameworks.} 3D models for object editing promise control over the geometry and viewpoint in 3D space. In Image Sculpting \cite{yenphraphai2024imagesculptingpreciseobject}, an object is reconstructed from an input image, modified in 3D and rendered back into 2D. While this offers geometric control, it is limited by the capabilities of current 3D reconstruction models \cite{voleti2024sv3d, liu2023zero1to3, shi2023zero123plus,hong2024lrmlargereconstructionmodel}, lacking accuracy for real-world applications and requiring significant manual effort and expertise. MVDrag3D \cite{chen2024mvdrag3d} conducts dragging-based 3D object editing through multi-view image generation. However, it does not allow user control of the object semantics or precise geometric constraints.
3D-DST \cite{ma2024generatingimages3dannotations} generates images with 3D annotations based on a class-specific 3D object. Their concept of utilizing a single 3D object in Blender to manage the viewpoint and then modify the objects' appearance in image space aligns with our approach.

\section{Methodology}
\label{sec:Methodology}
\begin{figure*}
    \begin{subfigure}{0.69\textwidth}
        \centering
        \includegraphics[width=0.97\linewidth]{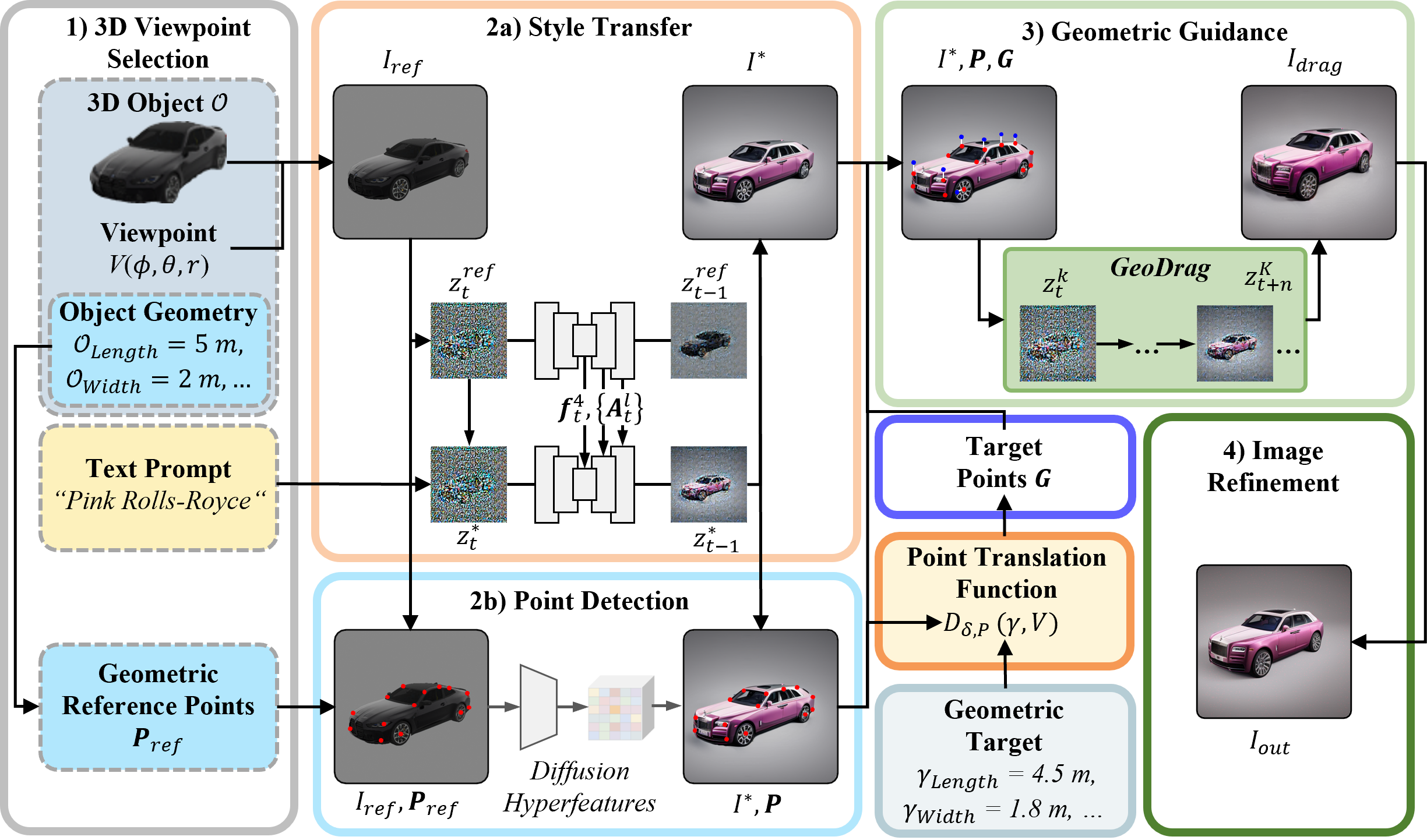}
        \caption{Architecture overview of the \textbf{GeoDiffusion} framework.}
        \label{fig:architecture}
    \end{subfigure}
    \begin{subfigure}{0.31\textwidth}
        \centering
        \includegraphics[width=0.97\linewidth]{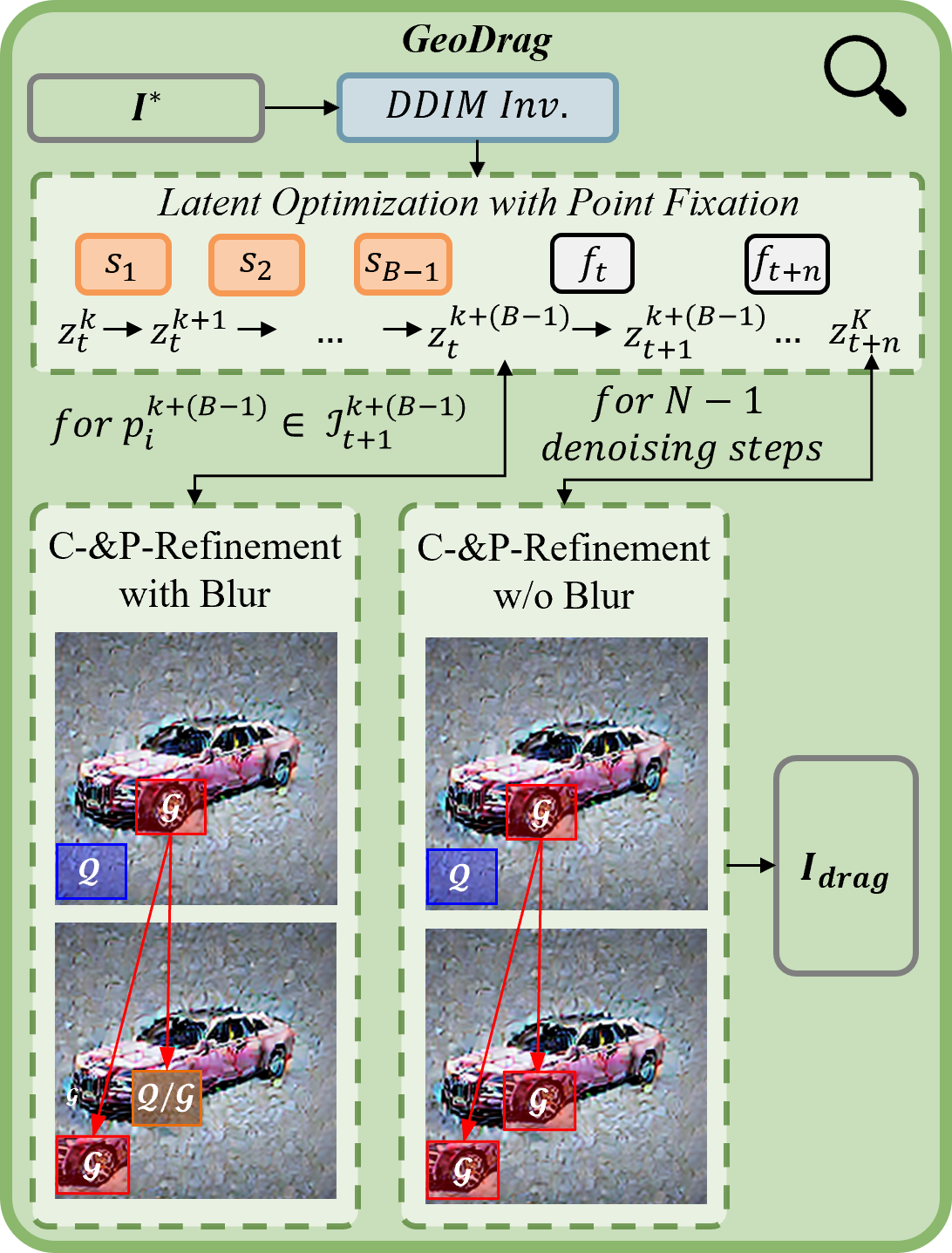}
        \caption{Architecture of \textbf{GeoDrag}.}
        \label{fig:Geodrag}
    \end{subfigure}
    \caption{(a) The GeoDiffusion architecture for accurate geometry guidance of objects in image space. A reference 3D object, its geometric features, a text prompt, and the geometric target are inputs for the framework. (b) For the dragging-based image modification, latent optimization with point fixation is done for $B$ iterations $g_k$ in each timestep $t$. At the end of each timestep, the patches $\mathcal{Q}$ around all points $p_i$ that are close to their target destination are copied and pasted to $\mathcal{G}$, and the initial patch is blurred. For the remaining denoising timesteps after the latent optimization, the patches $\mathcal{Q}$ are copied and pasted to $\mathcal{G}$ without blurring $\mathcal{Q}$. Best viewed when zoomed in.}
\end{figure*}

\begin{figure}
    \centering
    \includegraphics[width=0.96\linewidth]{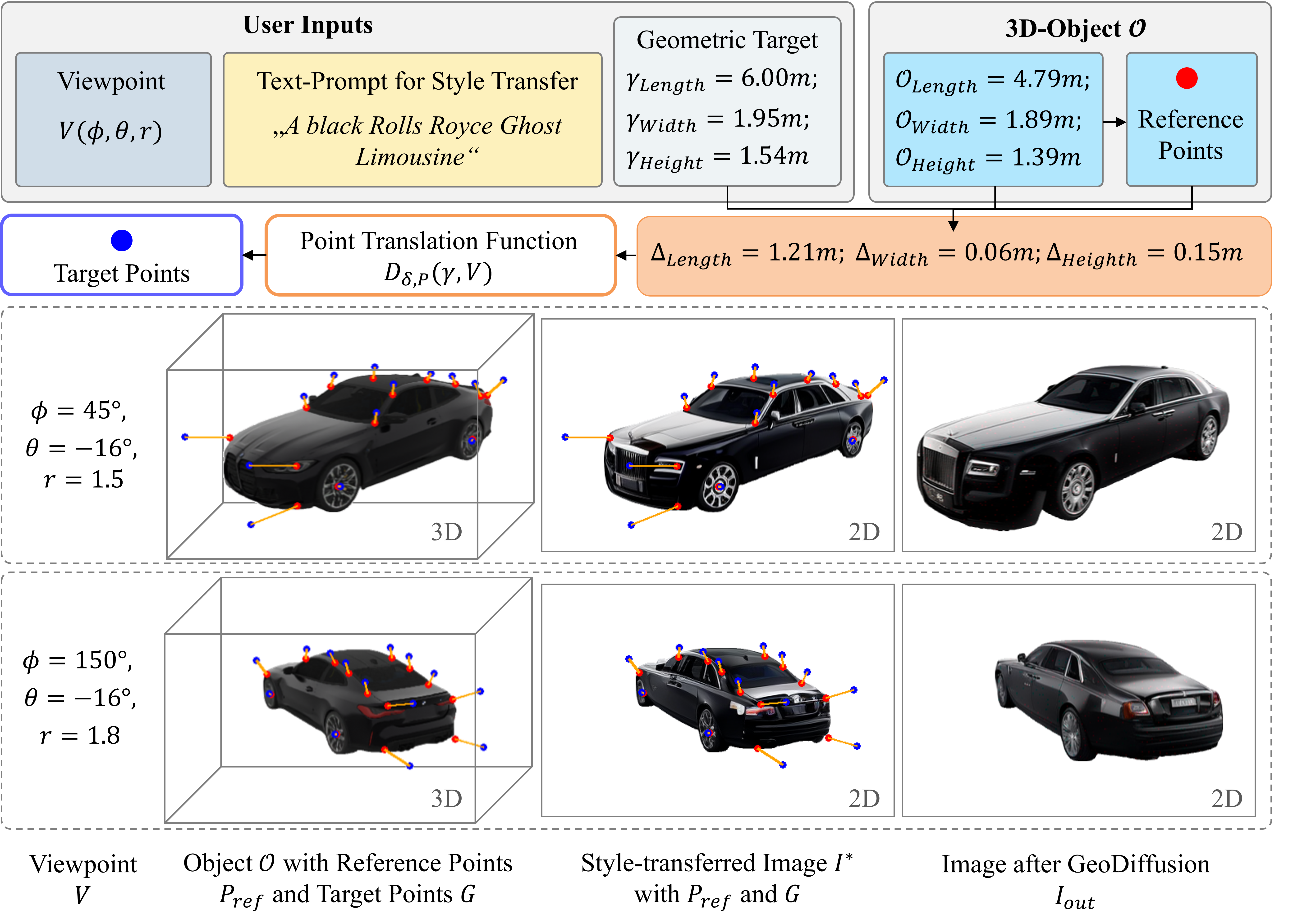}
    \vspace{-0.5em}
    \caption{\textbf{Geometry-guided generation of a car:} The user inputs the length, height, and width as the geometric target features. With the reference source points (red) and the point translation function, the target points (blue) are derived in 3D space. Using the viewpoint and the text prompt for style transfer, the image is rendered and subsequently modified in image space. Reference and target points are visualized in the 3D scene and the rendered image.}
    \label{fig:rebuttal_fig}
    \vspace{-1.5em}
\end{figure}

We address the core challenge of generating images that accurately adhere to precise 3D geometric constraints. Our four-step framework \textbf{GeoDiffusion} efficiently bridges the gap of conventional workflows, which either require extensive manual adjustments to correct for geometric distortions in 2D or demand resource-intensive modeling in 3D. Central to GeoDiffusion is a 3D-stage that ensures geometric consistency by explicitly encoding structural object features as reusable parametric constraints, guaranteeing accurate geometric control across varying viewpoints. Enforcing these changes directly in 2D would be difficult for two reasons: (a) Pure 2D image generation naturally struggles with viewpoint consistency. (b) Users have to manually compensate for viewpoint distortions that alter how modifications of 3D-object features appear in an image. The 3D-stage enables users to modify structural features without losing the proportions dictated by practical constraints (e.g., extending a car's length simultaneously affects its wheelbase and overhangs, which are both 3D features). \Cref{fig:rebuttal_fig} illustrates this process and shows how the same modification appears across different viewpoints and yet remains dimensionally correct with our framework. The architecture is visualized in~\Cref{fig:architecture} and consists of the following parts, which are explained in their own subsections of this chapter: 
\begin{enumerate}
    \item  We leverage a domain-specific reference 3D object to obtain its geometric characteristics as a baseline for the geometric modifications. This allows users to define source points based on the 3D reference geometry and specify how modifications (like increasing the wheelbase) propagate to each source point via the point translation function, thereby \textit{computing the target keypoints in 3D}. An image representation of a desired viewpoint is then created. Based on the image viewpoint information, both the source and the newly computed target points are automatically projected into the image domain.
    \item The obtained reference image is modified via image-to-image style transfer to fit the design target of the user. The source points in the restyled image have to be re-detected before conducting geometric image editing.
    \item The geometry of the object is modified through our improved dragging-based editing approach \textbf{GeoDrag}, specifically optimizing for dragging accuracy.
    \item The image is refined to increase visual quality, avoiding unwanted artifacts and distortions.
\end{enumerate}

\subsection{Viewpoint Selection in 3D}
\label{subsec:3D}
\noindent\textbf{Viewpoint Selection and Rendering.}
We opt to utilize a single, class-specific 3D reference object as a starting point. Domain-specific 3D objects (vehicles, buildings, furniture) can be obtained from publicly available or domain-specific datasets (ShapeNet \cite{chang2015shapenetinformationrich3dmodel}, Objaverse \cite{deitke2022objaverseuniverseannotated3d}, OmniObject3D \cite{wu2023omniobject3d}, DivAerNet \cite{elrefaie2024drivaernet}, 3DRealCar \cite{du20243drealcar}).
We load the 3D object into a Blender scene and define the desired position of the camera $\boldsymbol{c}$ through its distance to the object $r$, its elevation angle $\theta$ and the azimuth angle $\phi$. The synthesis of the reference image $I_{ref}$ that acts as input for the style transfer is defined as $I_{ref}=\mathcal{R}(\mathcal{O},\boldsymbol{c},\mathcal{K}, \mathcal{L})$, where $\mathcal{O}$ is the domain-specific 3D object, $\mathcal{R}$ is the rendering function and $\mathcal{K}$ and $\mathcal{L}$ are the camera intrinsic parameters and lighting conditions.
The source keypoints $P_{ref}$, describing the geometric features of the reference object, only need to be defined once per 3D reference object and can be reused across multiple geometric guidance tasks in the domain. This reuse allows for pre-implementation of standard geometric modifications and enables efficient experimentation with diverse constraints.

\noindent\textbf{Point Translation.}
To allow users to specify how the geometry can be modified, we propose a point translation function $\mathcal{D}$ that determines the logic of how the geometry is edited in 3D as a function of the displacement of the source keypoints $P_{ref} \in \mathbb{R}^{n\times3}$ to obtain the target points $G$.
It can formally be defined as $\mathcal{D}_{\delta,P_{ref}}(\gamma,V) \rightarrow G$, with $P_{ref},G \in \mathbb{R}^{n \times 3}$. $\gamma$ is the definable user input to modify the geometry (e.g., length, height, or wheelbase of a car; wingspan of an airplane), and $V$ is the viewpoint. $\delta$ defines the translation logic, describing how the points interact with regard to the user-defined parameters $\gamma$. 

\subsection{Style Transfer}
\label{subsec:Style_Trans}
\noindent\textbf{Image-to-Image Style Transfer.}
Transforming the rendered image representation $I_{ref}$ of the geometry-guided object via image-to-image style transfer is necessary to allow the user the generation of objects with desired styles and characteristics. Multiple style transfer methods enable the modification of images while preserving structural content \cite{hertz2022prompt,brooks2022instructpix2pix,wang2022pretraining,saharia2022paletteimagetoimagediffusionmodels}. We opt to use Plug-and-Play Diffusion \cite{tumanyan2022plugandplaydiffusionfeaturestextdriven} as it enables text-driven image-to-image translation without additional training, which supports the generalizability of the framework.

The objective is to generate an image $I^*$ that preserves the structure and the viewpoint of the object in the reference image $I_{ref}$ while reflecting the characteristics described by the user-prompt $W$. The reference image is inverted using DDIM~\cite{song2022denoisingdiffusionimplicitmodels} to obtain its initial noise $\boldsymbol{z}^{ref}_{T}$ (DDIM allows deterministic mapping between image and latent space). $\boldsymbol{x}^{ref}_{T}$ initializes the generation of the style-transferred image $I^*$ by setting the latent $\boldsymbol{z}^*_T=\boldsymbol{z}^{ref}_{T}$. For a timestep $t$, the guidance features from the fourth U-Net \cite{ronneberger2015unetconvolutionalnetworksbiomedical} layer $\{\boldsymbol{f}^4_t\}$ are extracted: $\boldsymbol{z}^{ref}_{t-1}=\boldsymbol{\epsilon}_{\theta}(\boldsymbol{z}^{ref}_T,\varnothing,t)$ and subsequently injected into the generation of $I^*$ by overriding the features $\{\boldsymbol{f}^{*4}_t\}$. The self-attention maps of higher resolution layers $\boldsymbol{A}^l_t$ are injected similarly, replacing those of the text-guided generation $\boldsymbol{A}^{*l}_t$. The operation to calculate the latent representation of the style-transferred image is denoted as \cite{tumanyan2022plugandplaydiffusionfeaturestextdriven}:
\begin{equation}
    \label{eq:PnP}
    \boldsymbol{z}^*_{t-1}=\boldsymbol{\hat{\epsilon}}_{\boldsymbol{\theta}}(\boldsymbol{z}^*_t,W,t;\boldsymbol{f}^4_t,\{ \boldsymbol{A}^l_t \}),
\end{equation}
where $\hat{\epsilon}_{\boldsymbol{\theta}}$ denotes the modified denoising function based on Stable Diffusion \cite{rombach2022highresolutionimagesynthesislatent}. We use $35$ DDIM sampling steps for the injection of $\boldsymbol{f}^4_t$ and $\boldsymbol{A}^l_t$.

\noindent\textbf{Reference-Point Detection.}
Since style transfer alters the object in the image, we need to detect the updated position of the source keypoints. There exist several approaches for this keypoint detection task \cite{dosovitskiy2021an,amir2021deep,zhang2023tale}. We use Diffusion Hyperfeatures \cite{luo2024diffusionhyperfeaturessearchingtime}. It extracts latent diffusion features over multiple timesteps and aggregates them into a single, interpretable Hyperfeature tensor. The reference points in $I^*$ are detected by finding the Hyperfeatures in the image with the maximum cosine similarity to the Hyperfeatures of the annotated points in the source image $I_{ref}$. 

\subsection{GeoDrag: Drag-based Geometry Guidance}
\noindent\textbf{Preliminary on Drag-based Editing.}
GeoDrag is inspired by GoodDrag \cite{zhang2024gooddraggoodpracticesdrag}, which achieves high accuracy (i.e. the dragged image features reliably reach the desired target position) and image fidelity. In GoodDrag, the editing operations are distributed across multiple steps within the diffusion process, alternating between dragging and denoising steps. $B$ dragging steps $s$ at time step $t$ are followed by a denoising step $f$. This rectifies perturbations introduced by the feature alignment when converting the latent image representation from $t$ to $t-1$. The feature alignment loss in GoodDrag \cite{zhang2024gooddraggoodpracticesdrag} is:
\begin{equation}
\label{eq:new_motion_loss}
\footnotesize
\begin{aligned}
\mathcal{L}(z_t^k; \{\boldsymbol{p}_i^k\}) = &\sum_{i=1}^l \left\Vert \mathrm{F}_{\mathrm{\Omega}(\boldsymbol{p}_i^k+\beta \boldsymbol{d}_i^k, r_1)}(z_t^k) - \text{sg}\left( \mathrm{F}_{\mathrm{\Omega}(\boldsymbol{p}_i^0, r_1)}(z_t^0) \right) \right\Vert_1 \\
&+ \lambda\left\Vert \left(z_{t-1}^k-{\text{sg}}\left(z_{t-1}^0\right)\right)\odot (1-\rm{M})\right\Vert_1,
\end{aligned}
\end{equation}
where $F_{\Omega}$ the features $F$ in the square patch around a point $p_i^k$.  $\boldsymbol{p}_i^0$ is the original handle point in the unedited image $z_t^0$. This formulation maintains consistency of the handle point $\boldsymbol{p}_i^k+\beta \boldsymbol{d}_i^k$ in the edited image $z_t^k$ with the original handle point, whose gradient is not calculated, as denoted by $sg(\cdot)$. $d_i^k$ is the movement direction of the point and $\beta$ its step size. The second term ensures that the non-editable region $1-M$ remains unaffected by the editing process.
The motion supervision for the $k$-th iteration at timestep $t$ takes a gradient descent step of step size $\eta$ according to the updated feature alignment loss $\mathcal{L}(z_t^k; \{\boldsymbol{p}_i^k\})$:
\begin{equation}
\label{eq:new_grad_descent}
\footnotesize
z_{t,j+1}^{k} = z_{t,j}^{k} - \eta \cdot \frac{\partial \mathcal{L}(z_{t,j}^{k}; \{\boldsymbol{p}_i^k\})}{\partial z_{t,j}^k}, ~~ j=0,\cdots,J-1.
\end{equation}
In this formulation, $z_{t,0}^{k} = z_{t}^{k}$ is the initial image, and $z_{t}^{k+1} = z_{t,J}^{k}$ is the output after $J$ gradient steps.

The necessary point-tracking to locate the new location of the handle point $\boldsymbol{p}_i^{k+1}$ is formulated as:
\begin{equation}
\label{eq:point_tracking}
\footnotesize
\boldsymbol{p}_i^{k+1} = \operatorname*{argmin}_{\boldsymbol{p} \in {\mathrm{\Omega}}(\boldsymbol{p}_i^k,r_2)} \left\Vert \mathrm{F}_{\boldsymbol{p}}(z_T^{k+1}) - \mathrm{F}_{\boldsymbol{p}_i^0}(z_T^0)\right\Vert_1.
\end{equation} 

\noindent\textbf{GeoDrag Method.}
The GoodDrag \cite{zhang2024gooddraggoodpracticesdrag} approach comes with certain limitations. Simultaneously optimizing multiple handle points causes them to influence each other adversely. Points that have already reached their target sometimes drift away in subsequent iterations due to the optimization of other points. This leads to inefficient convergence and negatively influences dragging accuracy, as there is no mechanism to fixate the points once they have reached their target location.

To address these limitations, we propose GeoDrag, providing the following improvements. We introduce a \textit{Point Fixation Mechanism} to prevent points from drifting away from their target location and employ a two-stage \textit{Copy-and-Paste Refinement} strategy inspired by SDE-Drag \cite{nie2024blessingrandomnesssdebeats}. The latter moves the latent image features that correspond to the points directly to their target destination to enhance precision and speed up convergence.\newline

\noindent \emph{Point Fixation Mechanism.} For defining the geometry points to fixate, we calculate the remaining distance of each point $i$ in iteration $k$ to its target destination as $e_i^k = \| g_i - {p}_i^k \|_2$. If the distance is smaller than the lower threshold $l$, the handle point $\boldsymbol{p}_i^k$ enters the pool of points ignored in the optimization, which is defined as $\mathcal{I}_t^k = \{{p}_i^k \ | \ e_i^k \leq l, \ i \in [n]\}$.
The motion supervision loss is formally modified by excluding the ignored points:
\begin{equation}
\label{eq:CustomDrag_EQ}
\scriptsize
\begin{aligned}
    \mathcal{L}(z_t^k; \{{p}_i^k\}) = &\sum_{i=1, \boldsymbol{p}_i^k \not\in \mathcal{I}_t^k}^n \left\| F_{\Omega({p}_i^k + \beta d_i^k, r_1)}(\hat{z}_t^k) - sg(F_{\Omega({p}_i^0, r_1)}(z_t)) \right\|_1 \\ &+ \ \lambda \left\| (\hat{z}_{t-1}^k - sg(z_{t-1})) \odot (1 - M_{\nabla}) \right\|_1 
\end{aligned}
\end{equation}\newline
To prevent ignored points from being affected by latent updates caused by the optimization of other points, we propose to set the gradient of the motion supervision loss to zero in regions around these points. We therefore define a gradient mask $M_{\nabla}$. Similar to the latent optimization, we do not apply the gradient mask to singular pixel positions but rather define a square patch $\Omega(p_i^k,r_{grad})$ with radius $r_{grad}$ around the corresponding point.
During the optimization, we continue to track the position of the points in $\mathcal{I}_t^k$. If a point in $\mathcal{I}_t^k$ drifts away from its position because of nearby feature patches of points being moved in the subsequent optimization and surpasses an upper threshold $e_i^k \geq u$, we remove it from the list of ignored points. Therefore, the gradient mask can be defined as:
\begin{equation}
\label{eq:Gradient_Mask}
\footnotesize
    M_{\nabla}[q] = 
        \begin{cases} 
        0, & \text{if } q \in \Omega(p_i^k, r_{grad}), \ p_i^k \in \mathcal{I}_t^k, \ e_i^k \leq l, \\
        0, & \text{if } q \in \Omega(p_i^k, r_{grad}), \ p_i^k \in \mathcal{I}_t^k, \ e_i^k \leq u, \\
        1, & \text{if } q \in \Omega(p_i^k, r_{grad}),  \ p_i^k \in \mathcal{I}_t^k, \ e_i^k \geq u. \\
        \end{cases} 
\end{equation}

\noindent \emph{Copy-and-Paste Refinement.} When a handle point is close to its target position, continuing the iterative optimization to cover the remaining distance becomes inefficient. To enhance the precision while improving efficiency, we propose to rather cover the small remaining distance to the target by directly copying and pasting the latent features.

At the end of each denoising timestep $t$, for each geometry point that is close to its target position $i$ in $\mathcal{I}_t^k$, we copy the latent features that correspond to the handle points' current position, defined by the square patch of radius $r_{cp}$, to the target position $g_i$:
\begin{equation} 
    z^k_t[\mathcal{G}] = \alpha z^k_t[\mathcal{Q}], \quad \text{with } \alpha \in (1,\infty).
\end{equation}
Here, $\mathcal{G} \in \Omega(g_i, r_{cp})$ and $\mathcal{Q} \in \Omega(p_i^k, r_{cp})$ define the square patches around the handle points $p_i^k$ and their target positions $g_i$ and $\alpha$ is an amplification factor. The notation in square brackets $z^k_t[\dots]$ denotes the latents $z^k_t$, defined by the patches $\mathcal{G}$ and $\mathcal{Q}$.
To avoid duplicate features in the image, latent features at the initial position of the handle point are blurred with Gaussian noise $\epsilon \sim \mathcal{N}(0, I)$, where $\mathcal{Q} \setminus \mathcal{G}$ denotes the blurring operation:
\begin{equation}
\footnotesize
    \begin{aligned}    
        z^k_t[\mathcal{Q} \setminus \mathcal{G}] = (\beta z^k_t + \sqrt{1 - \beta^2}\epsilon[\mathcal{Q} \setminus \mathcal{G}], \quad \text{with } \beta \in [0,1).
    \end{aligned}
\end{equation}
This process is repeated for all reference points once, at timestep $N$, after the optimization. There remains the possibility that not all points reach their target destination during the optimization process. To ensure that all reference points end up at the desired location in the image and stay there during the subsequent denoising process, we apply the copy-and-paste mechanism for the remaining $N-1$ denoising timesteps. Here, we directly copy the square patches $\mathcal{Q}$ with features from the original latent $z^0_t$ and the initial handle points ${p_i^0}$ and paste them to the target patch $\mathcal{G}$, obtaining $z_t[\mathcal{G}]=\alpha z_t^0[\mathcal{Q}]$.

\subsection{Image Refinement}
We incorporate image refinement in our approach to counter potential image perturbations and artifacts that can occur during dragging-based image editing. This enhances image fidelity and ensures that the output meets high standards for creative design applications. We opt for the refinement module shipped with Stable Diffusion XL \cite{podell2023sdxlimprovinglatentdiffusion}, passing the geometrically modified image $I_{drag}$ together with the text prompt $P$ to obtain the final image $I_{out}$. The choice of the guidance scale and strength is discussed in the appendix.

\section{Experiments}

\begin{figure*}
    \begin{subfigure}{0.54\textwidth}
        \centering
        \includegraphics[height=4.5cm]{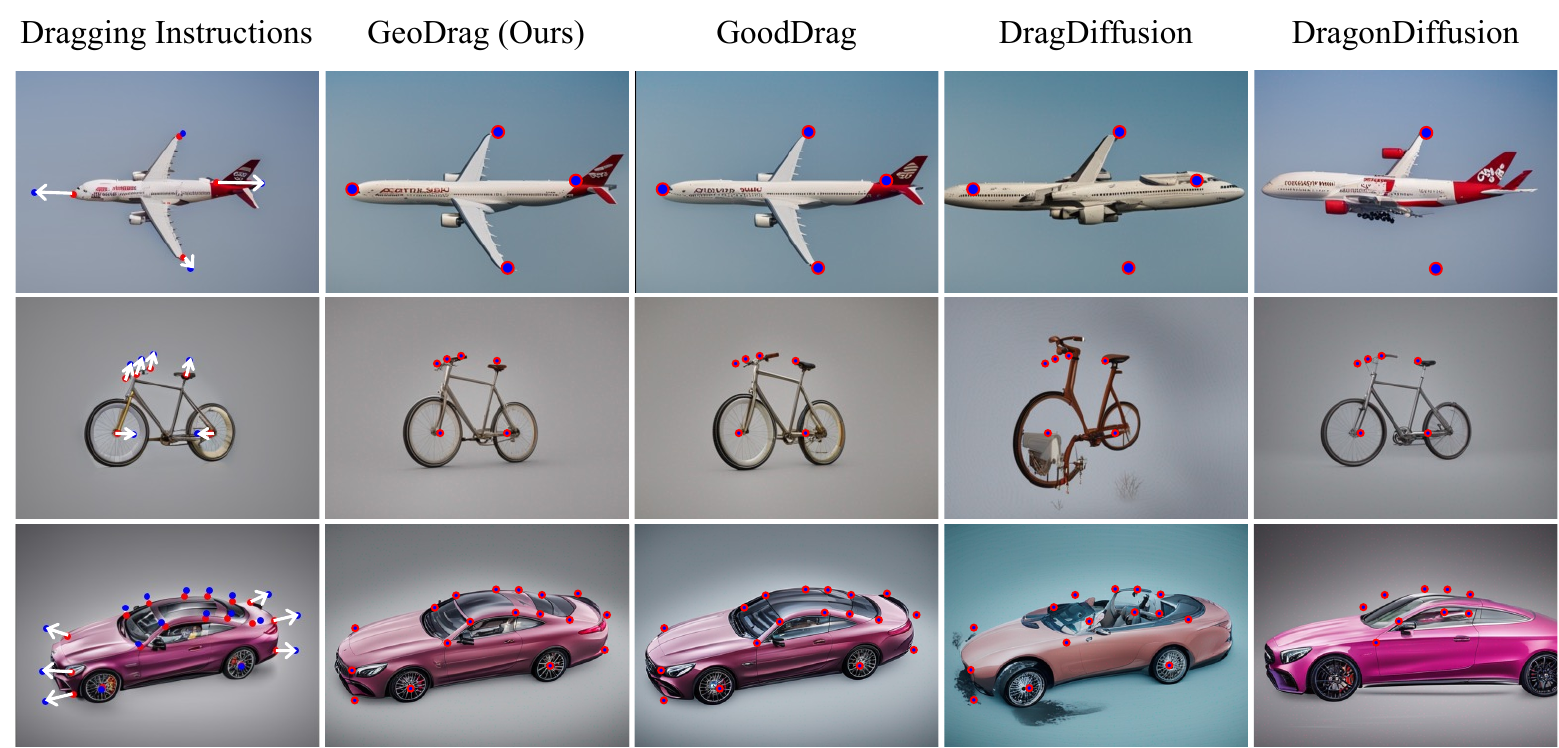}
        \caption{Qualitative Results of GeoDrag and other dragging methods on three examples from the geometry guidance dataset. The first column shows the dragging instructions. The target destinations of the points are annotated in each sample as red-blue dots. All tested methods were integrated into GeoDiffusion, and the dragging was conducted with the same initial images and instructions.}
        \label{fig:qualitative_examples}
    \end{subfigure}
    \hfill
    \begin{subfigure}{0.44\textwidth}
        \centering
        \includegraphics[height=4.5cm]{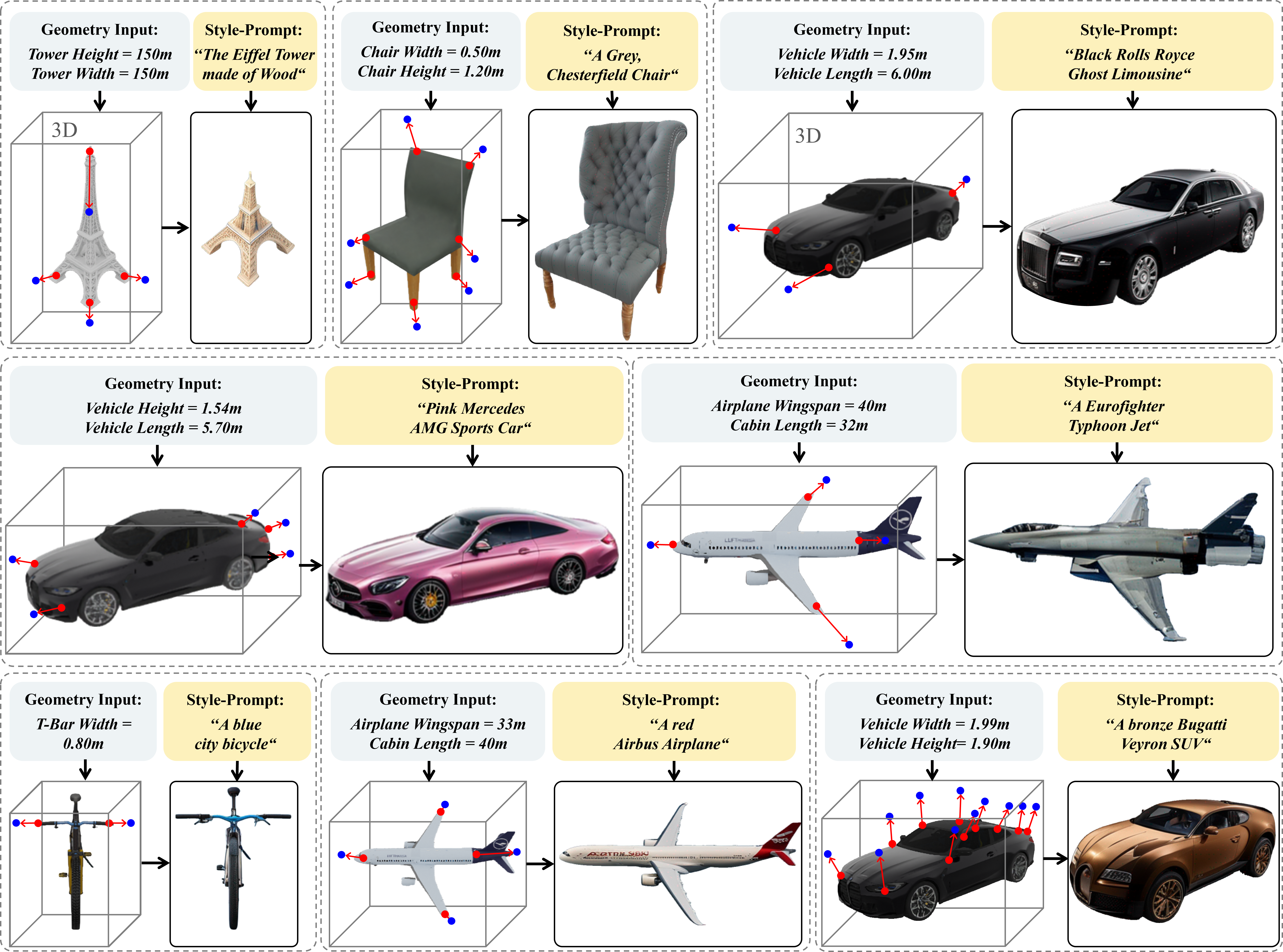}
        \caption{Qualitative results of GeoDiffusion with different objects. A class-specific 3D object (left) defines the target geometric characteristics of the object in 3D, which are projected as keypoint pairs. The geometric constraints are then projected into the 2D viewpoint, where, together with the style-prompt, the image is generated.}
        \label{fig:more_qualitative_examples}
    \end{subfigure}
    \vspace{-0.5em}
    \caption{Qualitative results of the GeoDiffusion framework. a) Comparison to other methods. b) Framework results.}
\end{figure*}

We evaluate the GeoDiffusion framework to verify its ability to generate images of objects with precise adherence to geometric constraints. Since our method is the first to introduce geometry-aware image generation conditioned on 3D priors, we begin with a qualitative assessment on multiple object categories (\Cref{subsec:GeoDiffEval}), ensuring the framework consistently applies 3D-informed generations across different domains.

To specifically assess the accuracy of geometric conditioning, we evaluate GeoDrag, the core mechanism responsible for modifying the image (\Cref{subsec:GeoDragEval}). We conduct a comparative analysis of different dragging methods by integrating alternative approaches into our pipeline and evaluating them under the same controlled conditions.

\subsection{Benchmark Datasets}
\label{subsec:Benchmark}
We define a geometry guidance benchmark dataset for guiding an object's geometric structure while maintaining its identity and viewpoint. The benchmark set contains 70 test cases, each consisting of a text prompt, a viewpoint, and geometric constraints. We test modifications on 40 cars, 18 bikes, and 12 airplane examples. We further assess our frameworks generation capabilities on 20 examples containing chairs and buildings.

For a general performance assessment of GeoDrag, we additionally use DragBench \cite{shi2023dragdiffusionharnessingdiffusionmodels}, which contains a diverse collection of images spanning various object categories, scenes, realistic and aesthetic styles.

\subsection{Evaluation of the GeoDiffusion Framework}
\label{subsec:GeoDiffEval}
\begin{figure}
    \centering
    \includegraphics[width=\linewidth]{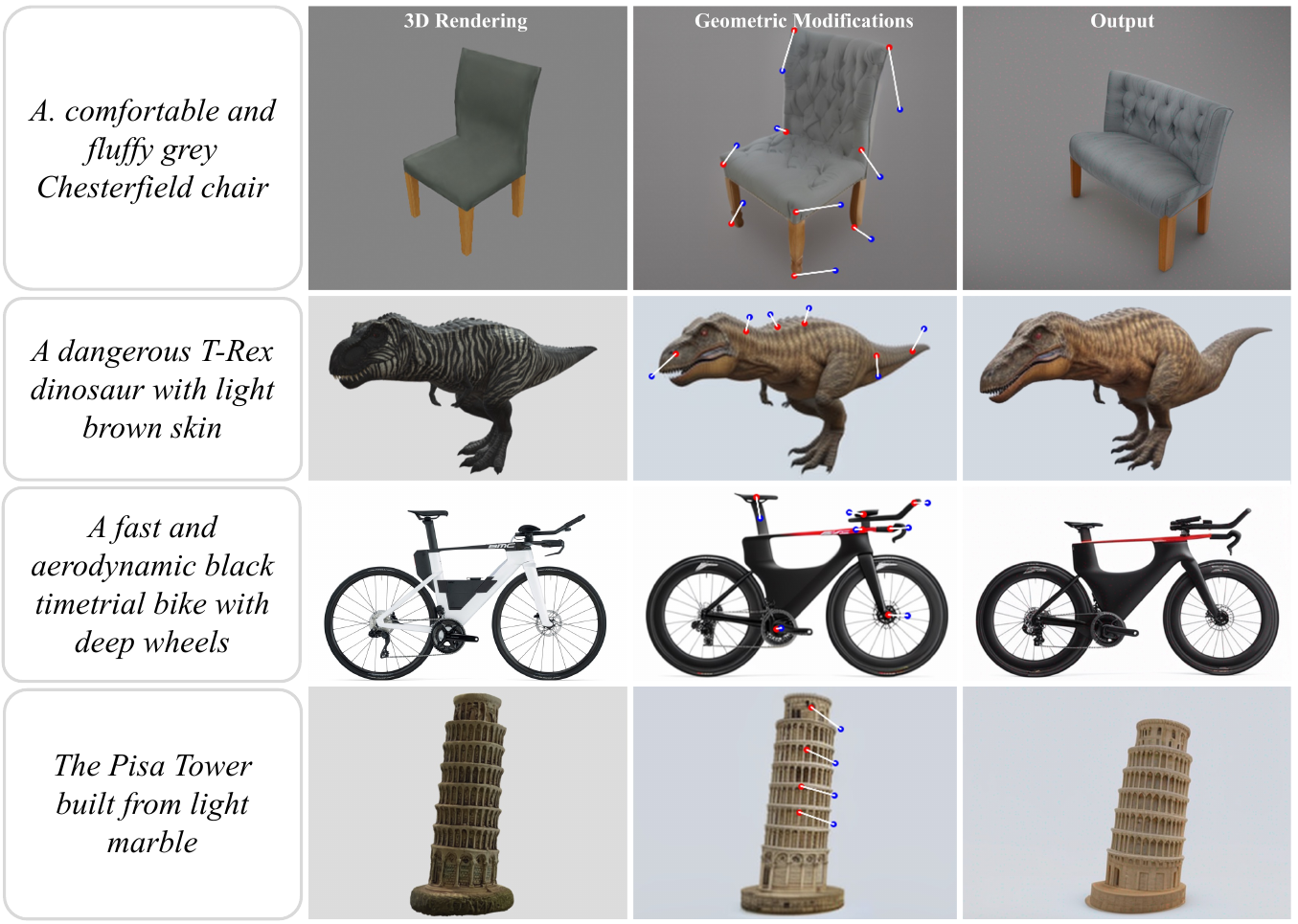}
    \vspace{-2em}
    \caption{GeoDiffusion with non-linear deformations.}
    \label{fig:Nonlinear_Examples}
    \vspace{-1em}
\end{figure}

We demonstrate the capability of GeoDiffusion to perform accurate geometric conditioning of 3D object features in 2D image generation. \Cref{fig:teaser} and \Cref{fig:more_qualitative_examples} visualize qualitative results on the geometry guidance benchmark. Our approach allows the modification of product design objects as well as other objects like chairs, buildings, bikes and animals (\Cref{fig:Nonlinear_Examples}).

Compared to other frameworks \cite{Mou_2024_CVPR, yenphraphai2024imagesculptingpreciseobject, chen2024mvdrag3d}, summarized in \Cref{tab:Comp_Table_Overall}, GeoDiffusion is unique in supporting 3D feature modification of objects in image space without the necessity to manually account for correct, viewpoint-dependent projections of the constraints. It is the only framework that enables the user to implement parametric relations to correlate the geometric keypoints. Therefore, this allows the introduction of design rules into the generation process.

\begin{table*}
\caption{High-level capability overview of frameworks for geometric feature modifications.}\vspace{-0.5em}
\scriptsize
\centering
\begin{tabularx}{\linewidth}{cccccc}
\toprule
Model & Custom Viewpoint Selection & Parametric Keypoint Relations & Editing in Image Space & 3D Object-Feature Modifications & Training-Free \\
\midrule
SDEdit \cite{meng2022sdedit} & \xmark & \xmark & \cmark & \xmark & \cmark	\\
DiffEditor \cite{Mou_2024_CVPR} & \xmark & \xmark & \xmark & (\cmark) & \xmark \\
MVDrag3D \cite{chen2024mvdrag3d} & \cmark & \xmark & \cmark & \cmark & \xmark \\
Image Sculpting \cite{yenphraphai2024imagesculptingpreciseobject} & \xmark & \xmark & \xmark & \cmark & \xmark \\
\midrule
\textbf{GeoDiffusion (ours)} & \cmark & \cmark & \cmark & \cmark & \cmark \\
\midrule
\bottomrule
\end{tabularx}
\label{tab:Comp_Table_Overall}
\end{table*}

\begin{figure}
    \centering
    \includegraphics[width=\linewidth]{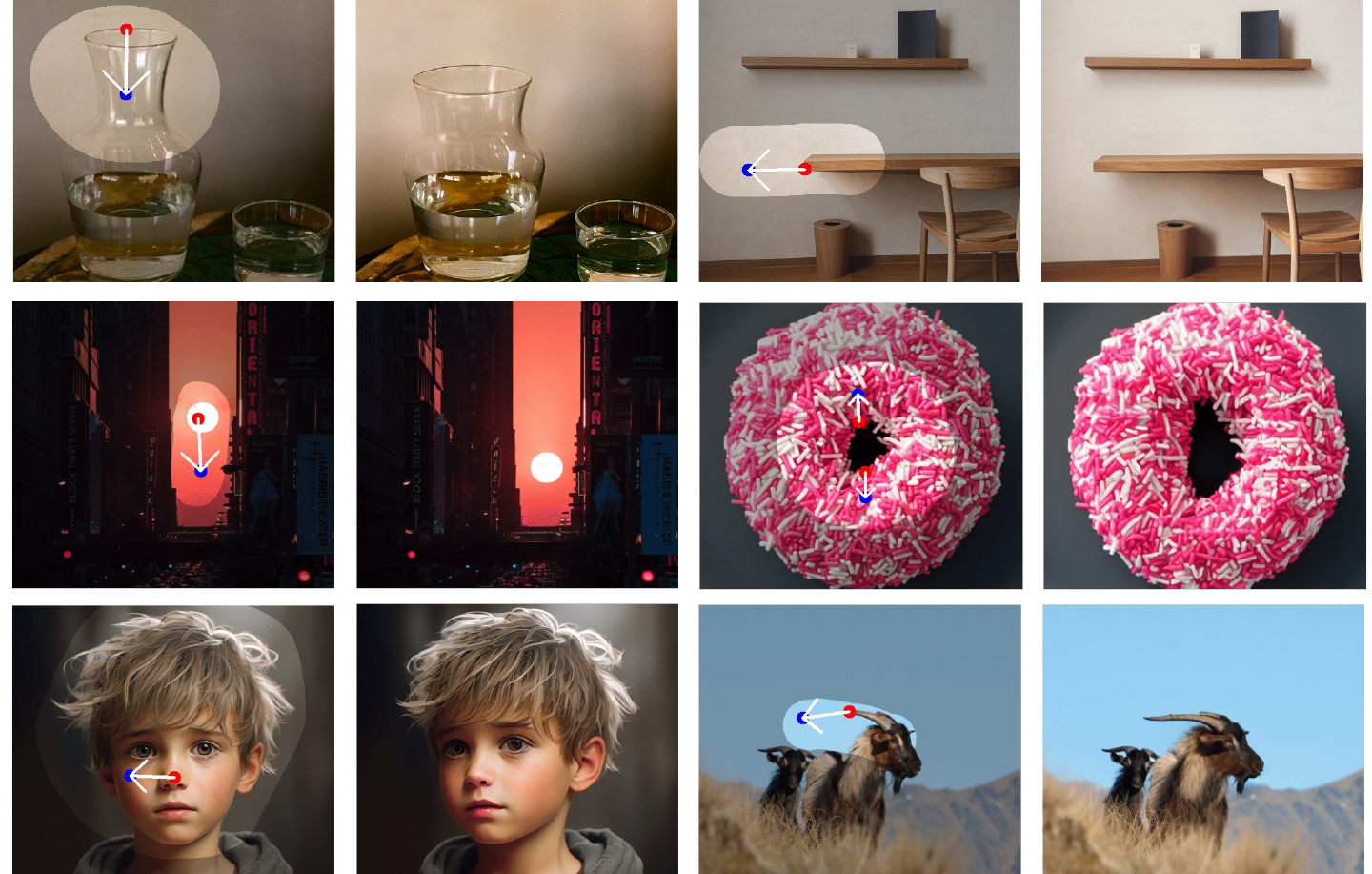}
    \vspace{-2em}
    \caption{Image modifications with GeoDrag on DragBench.}
    \label{fig:DragBench_examples}
    \vspace{-1em}
\end{figure}

\subsection{Evaluation of the GeoDrag Method}
\label{subsec:GeoDragEval}
\subsubsection{Qualitative Evaluation.} We specifically examine our improved dragging method, GeoDrag, against existing alternatives. For better comparability, we integrate the alternatives into our framework. \Cref{fig:qualitative_examples} visualizes examples from the geometry guidance dataset and \Cref{fig:DragBench_examples} shows modifications in traditional dragging tasks on DragBench. Compared to existing methods, GeoDrag achieves higher accuracy while maintaining object integrity. DragDiffusion \cite{shi2023dragdiffusionharnessingdiffusionmodels}, DragonDiffusion \cite{mou2023dragondiffusionenablingdragstylemanipulation}, FreeDrag \cite{ling2023freedragfeaturedraggingreliable}, DragNoise \cite{liu2024dragnoiseinteractivepointbased} and EasyDrag \cite{hou2024easydrag} struggle with this task. With GeoDrag, we can maintain the identity and structural integrity of modified objects in the image. The viewpoints and object semantics, such as color and design features, remain consistent. \Cref{fig:Nonlinear_Examples} shows some qualitative examples for nonlinear object modifications.
 
\subsubsection{Quantitative Evaluation} \label{subsec:Quantitative_Eval}
\noindent\textbf{Metrics.} For evaluating the accuracy of the geometric modifications, we calculate the mean distance (MD) between the reference points' position after dragging and their target position. The MD is calculated in pixels based on a $512\times512$ image.
We employ CLIP-score \cite{hessel-etal-2021-clipscore} to evaluate the alignment of the image with the given prompt.
To assess image fidelity and quality, we utilize HPSv2 \cite{wu2023humanpreferencescorev2}. 
We also measure the average execution time for the dragging. 
To be consistent with existing literature, we use the Image Fidelity (IF), defined as $1-\text{LPIPS}$ \cite{shi2023dragdiffusionharnessingdiffusionmodels, zhang2018unreasonableeffectivenessdeepfeatures}, for quantitative evaluation of the image quality on DragBench.

\noindent\textbf{Results.} \Cref{tab:evaluationdragging} summarizes the quantitative results on the geometry guidance and the DragBench datasets. GeoDrag achieves the highest accuracy of all evaluated approaches. The MD after the entire dragging process, including refinement, is $10.80$ pixels on average. This yields an $8.24\%$ increase in accuracy over the next best method, GoodDrag \cite{zhang2024gooddraggoodpracticesdrag} while simultaneously improving inference speed by $9.46\%$. In terms of CLIP and HPSv2 scores, GeoDrag is on par with GoodDrag. 

DragonDiffusion \cite{mou2023dragondiffusionenablingdragstylemanipulation} and SDE-Drag \cite{nie2024blessingrandomnesssdebeats} show higher inference speed but perform significantly worse in terms of accuracy. GeoDrag holds an advantage of $60.90\%$ over DragonDiffusion and  $54.02\%$ over SDE-Drag. We acknowledge that DragonDiffusion achieves higher quantitative scores for prompt alignment and image quality on the geometry guidance dataset. However, in addition to the deprecated accuracy of geometric modifications, we observe that it does not preserve the objects' integrity and viewpoint, seen in~\Cref{fig:qualitative_examples}. We notice a discrepancy between the quantitative HPSv2 metric and our visual examination. 

On DragBench, which contains mostly non-vehicle images, our approach consistently achieves the highest dragging accuracy among all tested methods. We increase the accuracy by $14.12\%$ compared to GoodDrag while being on par in terms of image fidelity. EasyDrag \cite{hou2024easydrag} achieves marginally better scores for image fidelity, but the mean distance of the dragged points is $11.69$ points higher. 

We provide an extended quantitative comparison of GeoDrag with GoodDrag in the appendix, showing that GeoDrag is on par with GoodDrag in terms of maintaining the style of the modified objects. The average difference in CLIP-similarity \cite{hessel-etal-2021-clipscore} is $0.38\%$, SSIM \cite{ssimwangImageQualityAssessment2004} differs by $0.15\%$, and LPIPS \cite{zhang2018unreasonableeffectivenessdeepfeatures} by $3.28\%$, measured between the style-transferred and final images. Both methods employ LoRA-based fine-tuning to maintain object identity. To prove the suitability of GeoDrag for modifying non-vehicle objects (beyond the DragBench cases shown in \Cref{tab:evaluationdragging}), we conduct modifications on 20 non-vehicle objects (chairs and buildings). GeoDrag outperforms GoodDrag by $32.7\%$ in dragging accuracy, achieves $4.1\%$ higher CLIP-scores, $2.2\%$ higher HPSv2-scores, and is $39.9\%$ faster, confirming that GeoDrag generalizes effectively to non-vehicle objects while preserving image fidelity.

\begin{table}
\caption{Evaluation of dragging methods on the Geometry Guidance Benchmark and DragBench. Best results are marked in bold.} \vspace{-0.5em}
    \centering
    \scriptsize
    \begin{tabularx}{\columnwidth}{YZZZZ|ZZ}
        \toprule 
        \multicolumn{1}{Y}{} & \multicolumn{4}{c}{\textbf{Geometry Guidance Dataset}} & \multicolumn{2}{c}{\textbf{DragBench}}\\
        \midrule
        \textbf{Method} &  \textbf{MD$\downarrow$} & \textbf{CLIP}$\uparrow$ &  \textbf{HPSv2}$\uparrow$ & \textbf{Time}$\downarrow$ & \textbf{MD}$\downarrow$ & \textbf{IF}$\uparrow$  \\
        \midrule
         DragDiffusion \cite{shi2023dragdiffusionharnessingdiffusionmodels}  & 44.38  & 23.09  & 22.61 & 50.53s  & 36.71 & 0.882 \\ 
         DragonDiffusion \cite{mou2023dragondiffusionenablingdragstylemanipulation}  & 27.62  & \textbf{26.83}  & \textbf{27.15} & \textbf{21.63s} & 52.21 & 0.899 \\ 
         SDEDrag \cite{nie2024blessingrandomnesssdebeats}  & 23.49  & 25.43  & 26.35 & 25.95s & 45.44 & 0.871 \\ 
         FreeDrag \cite{ling2023freedragfeaturedraggingreliable}  & 22.15  & 25.19  & 25.41 & 68.28s & 33.49 & 0.898 \\ 
         EasyDrag \cite{hou2024easydrag}  & 21.13  & 25.75  & 25.82 & 54.36s & 32.55 & \textbf{0.910} \\ 
         DragNoise \cite{liu2024dragnoiseinteractivepointbased} & 23.33  & 25.32  & 25.21 & 64.63s & 33.00 &  0.890 \\ 
         GoodDrag \cite{zhang2024gooddraggoodpracticesdrag} & 11.77  & 25.21  & 25.93 & 66.30s & 24.29 & 0.871 \\
         \midrule
         \textbf{GeoDrag}  & \textbf{10.80}  & 25.31  & 25.71 & 60.03s & \textbf{20.86} & 0.872 \\
         \bottomrule
    \end{tabularx}
\label{tab:evaluationdragging}
\vspace{-1em}
\end{table}

\subsection{Ablations}
\label{subsec:Ablations}
\textbf{Hyperparameter Configurations.} 
We test different GeoDrag configurations to evaluate the trade-off between dragging accuracy and inference speed. The results show that setting the thresholds $l$ and $u$ for the point fixation closer to the target destination of the points increases accuracy at the cost of speed. This is intuitive as the points remain in the iterative optimization for longer and re-enter it earlier in case of being dragged away. Additionally, we observe a decrease in dragging accuracy by $16\%$ if the points in $\mathcal{I}$ are not re-entered into the iterative optimization in case of being dragged away beyond $u$.

Running the dragging for more steps increases the duration but results in better accuracy. Of the tested configurations that obtain acceptable accuracy, we achieve a mean duration for image modification of $40.77s$, which poses a $38.5\%$ increase in inference speed over GoodDrag. The fastest configuration, which is on par with GoodDrag in terms of accuracy, still achieves a $25.7\%$ speed increase. 

\noindent\textbf{Copy-and-Paste Refinement.} When the copy-and-paste refinement that follows the iterative optimization is not carried out, the MD increases by $2.73$ pixels on the geometry guidance dataset. Skipping the refinement after each iterative optimization step decreases accuracy by $0.19$ pixels.

\noindent\textbf{Image Refinement.} Refining the images with SDXL \cite{podell2023sdxlimprovinglatentdiffusion} after the geometric modification results in a trade-off. On the geometry guidance dataset, it increases the MD by $1.45$ pixels from $9.35$ to $10.8$ but improves image quality noticeably as the HPSv2-score increases by $1.49$ points. This is also confirmed by visual examination.

\subsection{Limitations} 
GeoDiffusion fundamentally depends on the capabilities of the pre-trained diffusion models, which can impact results in domain-specific cases. In cases of large dragging distances, preserving precise geometry can be challenging. Consistency across multiple viewpoints remains an open area for improvement, as each generation is currently an independent process, leading to variations in style and color.
\section{Conclusion}
With this work, we introduce GeoDiffusion, a training-free framework for precise geometric control of 3D object features in image generation. By utilizing a single, class-specific 3D object, we eliminate the need for extensive training, enabling users to define geometric keypoints and parametric relationships in 3D space. To ensure viewpoint consistency, we first render the desired viewpoint and then perform text-conditioned image-to-image style transfer. Additionally, we propose GeoDrag, which enhances dragging accuracy and achieves better results for geometric modifications, as well as on the DragBench benchmark. 

\noindent\textbf{Future Research} could aim to improve image-to-image translation for enhanced alignment with user intentions and higher image fidelity. Another direction is to incorporate GeoDiffusion within a multi-view image model, advancing toward consistent 3D object modifications without the need for complex manipulations in a 3D environment.

{
    \small
    \bibliographystyle{ieeenat_fullname}
    \bibliography{main}

\begin{thebibliography}{52}
\providecommand{\natexlab}[1]{#1}
\providecommand{\url}[1]{\texttt{#1}}
\expandafter\ifx\csname urlstyle\endcsname\relax
  \providecommand{\doi}[1]{doi: #1}\else
  \providecommand{\doi}{doi: \begingroup \urlstyle{rm}\Url}\fi

\bibitem[Adobe(2024{\natexlab{a}})]{AdobeAfterEffects}
Adobe.
\newblock After effects, 2024{\natexlab{a}}.

\bibitem[Adobe(2024{\natexlab{b}})]{AdobePhotoshop}
Adobe.
\newblock Photoshop, 2024{\natexlab{b}}.

\bibitem[Amir et~al.(2022)Amir, Gandelsman, Bagon, and Dekel]{amir2021deep}
Shir Amir, Yossi Gandelsman, Shai Bagon, and Tali Dekel.
\newblock Deep vit features as dense visual descriptors.
\newblock \emph{ECCVW What is Motion For?}, 2022.

\bibitem[Blender(2024)]{Blender}
Blender.
\newblock Blender, 2024.

\bibitem[Brooks et~al.(2023)Brooks, Holynski, and Efros]{brooks2022instructpix2pix}
Tim Brooks, Aleksander Holynski, and Alexei~A. Efros.
\newblock Instructpix2pix: Learning to follow image editing instructions.
\newblock In \emph{CVPR}, 2023.

\bibitem[Chang et~al.(2015)Chang, Funkhouser, Guibas, Hanrahan, Huang, Li, Savarese, Savva, Song, Su, Xiao, Yi, and Yu]{chang2015shapenetinformationrich3dmodel}
Angel~X. Chang, Thomas Funkhouser, Leonidas Guibas, Pat Hanrahan, Qixing Huang, Zimo Li, Silvio Savarese, Manolis Savva, Shuran Song, Hao Su, Jianxiong Xiao, Li Yi, and Fisher Yu.
\newblock {ShapeNet: An Information-Rich 3D Model Repository}.
\newblock Technical Report arXiv:1512.03012 [cs.GR], Stanford University --- Princeton University --- Toyota Technological Institute at Chicago, 2015.

\bibitem[Chen et~al.(2024)Chen, Lan, Chen, Zhou, and Pan]{chen2024mvdrag3d}
Honghua Chen, Yushi Lan, Yongwei Chen, Yifan Zhou, and Xingang Pan.
\newblock Mvdrag3d: Drag-based creative 3d editing via multi-view generation-reconstruction priors.
\newblock \emph{arXiv preprint arXiv:2410.16272}, 2024.

\bibitem[Deitke et~al.(2023)Deitke, Schwenk, Salvador, Weihs, Michel, VanderBilt, Schmidt, Ehsanit, Kembhavi, and Farhadi]{deitke2022objaverseuniverseannotated3d}
Matt Deitke, Dustin Schwenk, Jordi Salvador, Luca Weihs, Oscar Michel, Eli VanderBilt, Ludwig Schmidt, Kiana Ehsanit, Aniruddha Kembhavi, and Ali Farhadi.
\newblock { Objaverse: A Universe of Annotated 3D Objects }.
\newblock In \emph{2023 IEEE/CVF Conference on Computer Vision and Pattern Recognition (CVPR)}, pages 13142--13153, Los Alamitos, CA, USA, 2023. IEEE Computer Society.

\bibitem[Dhariwal and Nichol(2021)]{dhariwal2021diffusionmodelsbeatgans}
Prafulla Dhariwal and Alex Nichol.
\newblock Diffusion models beat gans on image synthesis.
\newblock In \emph{Proceedings of the 35th International Conference on Neural Information Processing Systems}, Red Hook, NY, USA, 2021. Curran Associates Inc.

\bibitem[Dosovitskiy et~al.(2021)Dosovitskiy, Beyer, Kolesnikov, Weissenborn, Zhai, Unterthiner, Dehghani, Minderer, Heigold, Gelly, Uszkoreit, and Houlsby]{dosovitskiy2021an}
Alexey Dosovitskiy, Lucas Beyer, Alexander Kolesnikov, Dirk Weissenborn, Xiaohua Zhai, Thomas Unterthiner, Mostafa Dehghani, Matthias Minderer, Georg Heigold, Sylvain Gelly, Jakob Uszkoreit, and Neil Houlsby.
\newblock An image is worth 16x16 words: Transformers for image recognition at scale.
\newblock In \emph{International Conference on Learning Representations}, 2021.

\bibitem[Du et~al.(2024)Du, Sun, Wang, Wu, Sheng, Ying, Lu, Zhu, Zhan, and Yu]{du20243drealcar}
Xiaobiao Du, Haiyang Sun, Shuyun Wang, Zhuojie Wu, Hongwei Sheng, Jiaying Ying, Ming Lu, Tianqing Zhu, Kun Zhan, and Xin Yu.
\newblock 3drealcar: An in-the-wild rgb-d car dataset with 360-degree views.
\newblock \emph{arXiv preprint arXiv:2406.04875}, 2024.

\bibitem[Elrefaie et~al.(2024)Elrefaie, Dai, and Ahmed]{elrefaie2024drivaernet}
Mohamed Elrefaie, Angela Dai, and Faez Ahmed.
\newblock Drivaernet: A parametric car dataset for data-driven aerodynamic design and graph-based drag prediction.
\newblock \emph{arXiv preprint arXiv:2403.08055}, 2024.

\bibitem[Hertz et~al.(2023)Hertz, Mokady, Tenenbaum, Aberman, Pritch, and Cohen-Or]{hertz2022prompt}
Amir Hertz, Ron Mokady, Jay Tenenbaum, Kfir Aberman, Yael Pritch, and Daniel Cohen-Or.
\newblock Prompt-to-prompt image editing with cross attention control.
\newblock In \emph{ICLR}, 2023.

\bibitem[Hessel et~al.(2021)Hessel, Holtzman, Forbes, Le~Bras, and Choi]{hessel-etal-2021-clipscore}
Jack Hessel, Ari Holtzman, Maxwell Forbes, Ronan Le~Bras, and Yejin Choi.
\newblock {CLIPS}core: A reference-free evaluation metric for image captioning.
\newblock In \emph{Proceedings of the 2021 Conference on Empirical Methods in Natural Language Processing}, pages 7514--7528, 2021.

\bibitem[Ho et~al.(2020)Ho, Jain, and Abbeel]{ho2020denoisingdiffusionprobabilisticmodels}
Jonathan Ho, Ajay Jain, and Pieter Abbeel.
\newblock Denoising diffusion probabilistic models.
\newblock In \emph{Advances in Neural Information Processing Systems}, pages 6840--6851. Curran Associates, Inc., 2020.

\bibitem[Hong et~al.(2024)Hong, Zhang, Gu, Bi, Zhou, Liu, Liu, Sunkavalli, Bui, and Tan]{hong2024lrmlargereconstructionmodel}
Yicong Hong, Kai Zhang, Jiuxiang Gu, Sai Bi, Yang Zhou, Difan Liu, Feng Liu, Kalyan Sunkavalli, Trung Bui, and Hao Tan.
\newblock Lrm: Large reconstruction model for single image to 3d, 2024.

\bibitem[Hou et~al.(2024)Hou, Liu, Zhang, Liu, Liu, and You]{hou2024easydrag}
Xingzhong Hou, Boxiao Liu, Yi Zhang, Jihao Liu, Yu Liu, and Haihang You.
\newblock Easydrag: Efficient point-based manipulation on diffusion models.
\newblock In \emph{Proceedings of the IEEE/CVF Conference on Computer Vision and Pattern Recognition}, pages 8404--8413, 2024.

\bibitem[Hu et~al.(2022)Hu, yelong shen, Wallis, Allen-Zhu, Li, Wang, Wang, and Chen]{hu2021loralowrankadaptationlarge}
Edward~J Hu, yelong shen, Phillip Wallis, Zeyuan Allen-Zhu, Yuanzhi Li, Shean Wang, Lu Wang, and Weizhu Chen.
\newblock Lo{RA}: Low-rank adaptation of large language models.
\newblock In \emph{International Conference on Learning Representations}, 2022.

\bibitem[Kawar et~al.(2023)Kawar, Zada, Lang, Tov, Chang, Dekel, Mosseri, and Irani]{kawar2023imagic}
Bahjat Kawar, Shiran Zada, Oran Lang, Omer Tov, Huiwen Chang, Tali Dekel, Inbar Mosseri, and Michal Irani.
\newblock Imagic: Text-based real image editing with diffusion models.
\newblock In \emph{Conference on Computer Vision and Pattern Recognition 2023}, 2023.

\bibitem[Kollovieh et~al.(2023)Kollovieh, Ansari, Bohlke-Schneider, Zschiegner, Wang, and Wang]{kollovieh2023predict}
Marcel Kollovieh, Abdul~Fatir Ansari, Michael Bohlke-Schneider, Jasper Zschiegner, Hao Wang, and Yuyang~Bernie Wang.
\newblock Predict, refine, synthesize: Self-guiding diffusion models for probabilistic time series forecasting.
\newblock \emph{Advances in Neural Information Processing Systems}, 36:\penalty0 28341--28364, 2023.

\bibitem[Ling et~al.(2024)Ling, Chen, Zhang, Chen, Jin, and Zheng]{ling2023freedragfeaturedraggingreliable}
Pengyang Ling, Lin Chen, Pan Zhang, Huaian Chen, Yi Jin, and Jinjin Zheng.
\newblock Freedrag: Feature dragging for reliable point-based image editing.
\newblock In \emph{Proceedings of the IEEE/CVF Conference on Computer Vision and Pattern Recognition}, pages 6860--6870, 2024.

\bibitem[Liu et~al.(2024)Liu, Xu, Yang, Zeng, and He]{liu2024dragnoiseinteractivepointbased}
Haofeng Liu, Chenshu Xu, Yifei Yang, Lihua Zeng, and Shengfeng He.
\newblock Drag your noise: Interactive point-based editing via diffusion semantic propagation.
\newblock \emph{arXiv preprint 2404.01050}, 2024.

\bibitem[Liu et~al.(2023)Liu, Wu, Hoorick, Tokmakov, Zakharov, and Vondrick]{liu2023zero1to3}
Ruoshi Liu, Rundi Wu, Basile~Van Hoorick, Pavel Tokmakov, Sergey Zakharov, and Carl Vondrick.
\newblock Zero-1-to-3: Zero-shot one image to 3d object.
\newblock \emph{arxiv preprint 2303.11328}, 2023.

\bibitem[Lugmayr et~al.(2022)Lugmayr, Danelljan, Romero, Yu, Timofte, and Van~Gool]{Lugmayr_2022_CVPR}
Andreas Lugmayr, Martin Danelljan, Andres Romero, Fisher Yu, Radu Timofte, and Luc Van~Gool.
\newblock Repaint: Inpainting using denoising diffusion probabilistic models.
\newblock In \emph{Proceedings of the IEEE/CVF Conference on Computer Vision and Pattern Recognition (CVPR)}, pages 11461--11471, 2022.

\bibitem[Luo et~al.(2023)Luo, Dunlap, Park, Holynski, and Darrell]{luo2024diffusionhyperfeaturessearchingtime}
Grace Luo, Lisa Dunlap, Dong~Huk Park, Aleksander Holynski, and Trevor Darrell.
\newblock Diffusion hyperfeatures: Searching through time and space for semantic correspondence.
\newblock In \emph{Advances in Neural Information Processing Systems}, 2023.

\bibitem[Ma et~al.(2024)Ma, Liu, Wang, Wang, Yuan, Zhang, Xiao, Zhang, Lu, Duan, Qi, Kortylewski, Liu, and Yuille]{ma2024generatingimages3dannotations}
Wufei Ma, Qihao Liu, Jiahao Wang, Angtian Wang, Xiaoding Yuan, Yi Zhang, Zihao Xiao, Guofeng Zhang, Beijia Lu, Ruxiao Duan, Yongrui Qi, Adam Kortylewski, Yaoyao Liu, and Alan Yuille.
\newblock Generating images with 3d annotations using diffusion models.
\newblock In \emph{The Twelfth International Conference on Learning Representations}, 2024.

\bibitem[Meng et~al.(2022)Meng, He, Song, Song, Wu, Zhu, and Ermon]{meng2022sdedit}
Chenlin Meng, Yutong He, Yang Song, Jiaming Song, Jiajun Wu, Jun-Yan Zhu, and Stefano Ermon.
\newblock {SDE}dit: Guided image synthesis and editing with stochastic differential equations.
\newblock In \emph{International Conference on Learning Representations}, 2022.

\bibitem[Mou et~al.(2023)Mou, Wang, Xie, Wu, Zhang, Qi, Shan, and Qie]{mou2023t2iadapterlearningadaptersdig}
Chong Mou, Xintao Wang, Liangbin Xie, Yanze Wu, Jian Zhang, Zhongang Qi, Ying Shan, and Xiaohu Qie.
\newblock T2i-adapter: Learning adapters to dig out more controllable ability for text-to-image diffusion models.
\newblock \emph{arXiv preprint arXiv:2302.08453}, 2023.

\bibitem[Mou et~al.(2024{\natexlab{a}})Mou, Wang, Song, Shan, and Zhang]{Mou_2024_CVPR}
Chong Mou, Xintao Wang, Jiechong Song, Ying Shan, and Jian Zhang.
\newblock Diffeditor: Boosting accuracy and flexibility on diffusion-based image editing.
\newblock In \emph{Proceedings of the IEEE/CVF Conference on Computer Vision and Pattern Recognition (CVPR)}, pages 8488--8497, 2024{\natexlab{a}}.

\bibitem[Mou et~al.(2024{\natexlab{b}})Mou, Wang, Song, Shan, and Zhang]{mou2023dragondiffusionenablingdragstylemanipulation}
Chong Mou, Xintao Wang, Jiechong Song, Ying Shan, and Jian Zhang.
\newblock Dragondiffusion: Enabling drag-style manipulation on diffusion models.
\newblock In \emph{The Twelfth International Conference on Learning Representations}, 2024{\natexlab{b}}.

\bibitem[Nie et~al.(2024)Nie, Guo, Lu, Zhou, Zheng, and Li]{nie2024blessingrandomnesssdebeats}
Shen Nie, Hanzhong~Allan Guo, Cheng Lu, Yuhao Zhou, Chenyu Zheng, and Chongxuan Li.
\newblock The blessing of randomness: {SDE} beats {ODE} in general diffusion-based image editing.
\newblock In \emph{The Twelfth International Conference on Learning Representations}, 2024.

\bibitem[Pan et~al.(2023)Pan, Tewari, Leimkuehler, Liu, Meka, and Theobalt]{pan2023dragganinteractivepointbased}
Xingang Pan, Ayush Tewari, Thomas Leimkuehler, Lingjie Liu, Abhimitra Meka, and Christian Theobalt.
\newblock Drag your gan: Interactive point-based manipulation on the generative image manifold.
\newblock In \emph{ACM SIGGRAPH 2023 Conference Proceedings}, 2023.

\bibitem[Podell et~al.(2024)Podell, English, Lacey, Blattmann, Dockhorn, M{\"u}ller, Penna, and Rombach]{podell2023sdxlimprovinglatentdiffusion}
Dustin Podell, Zion English, Kyle Lacey, Andreas Blattmann, Tim Dockhorn, Jonas M{\"u}ller, Joe Penna, and Robin Rombach.
\newblock {SDXL}: Improving latent diffusion models for high-resolution image synthesis.
\newblock In \emph{The Twelfth International Conference on Learning Representations}, 2024.

\bibitem[Rombach et~al.(2022)Rombach, Blattmann, Lorenz, Esser, and Ommer]{rombach2022highresolutionimagesynthesislatent}
Robin Rombach, Andreas Blattmann, Dominik Lorenz, Patrick Esser, and Bjorn Ommer.
\newblock { High-Resolution Image Synthesis with Latent Diffusion Models }.
\newblock In \emph{2022 IEEE/CVF Conference on Computer Vision and Pattern Recognition (CVPR)}, pages 10674--10685, Los Alamitos, CA, USA, 2022. IEEE Computer Society.

\bibitem[Ronneberger et~al.(2015)Ronneberger, Fischer, and Brox]{ronneberger2015unetconvolutionalnetworksbiomedical}
Olaf Ronneberger, Philipp Fischer, and Thomas Brox.
\newblock U-net: Convolutional networks for biomedical image segmentation.
\newblock In \emph{LNCS}, pages 234--241, 2015.

\bibitem[Saharia et~al.(2022)Saharia, Chan, Chang, Lee, Ho, Salimans, Fleet, and Norouzi]{saharia2022paletteimagetoimagediffusionmodels}
Chitwan Saharia, William Chan, Huiwen Chang, Chris Lee, Jonathan Ho, Tim Salimans, David Fleet, and Mohammad Norouzi.
\newblock Palette: Image-to-image diffusion models.
\newblock In \emph{ACM SIGGRAPH 2022 Conference Proceedings}, New York, NY, USA, 2022. Association for Computing Machinery.

\bibitem[Shi et~al.(2023{\natexlab{a}})Shi, Chen, Zhang, Liu, Xu, Wei, Chen, Zeng, and Su]{shi2023zero123plus}
Ruoxi Shi, Hansheng Chen, Zhuoyang Zhang, Minghua Liu, Chao Xu, Xinyue Wei, Linghao Chen, Chong Zeng, and Hao Su.
\newblock Zero123++: a single image to consistent multi-view diffusion base model.
\newblock \emph{arxiv preprint arXiv:2310.15110}, 2023{\natexlab{a}}.

\bibitem[Shi et~al.(2023{\natexlab{b}})Shi, Xue, Pan, Zhang, Tan, and Bai]{shi2023dragdiffusionharnessingdiffusionmodels}
Yujun Shi, Chuhui Xue, Jiachun Pan, Wenqing Zhang, Vincent~YF Tan, and Song Bai.
\newblock Dragdiffusion: Harnessing diffusion models for interactive point-based image editing.
\newblock \emph{arXiv preprint 2306.14435}, 2023{\natexlab{b}}.

\bibitem[Sohl-Dickstein et~al.(2015)Sohl-Dickstein, Weiss, Maheswaranathan, and Ganguli]{sohldickstein2015deepunsupervisedlearningusing}
Jascha Sohl-Dickstein, Eric Weiss, Niru Maheswaranathan, and Surya Ganguli.
\newblock Deep unsupervised learning using nonequilibrium thermodynamics.
\newblock In \emph{Proceedings of the 32nd International Conference on Machine Learning}, pages 2256--2265, Lille, France, 2015. PMLR.

\bibitem[Song et~al.(2020)Song, Meng, and Ermon]{song2022denoisingdiffusionimplicitmodels}
Jiaming Song, Chenlin Meng, and Stefano Ermon.
\newblock Denoising diffusion implicit models.
\newblock \emph{arXiv:2010.02502}, 2020.

\bibitem[Tumanyan et~al.(2023)Tumanyan, Geyer, Bagon, and Dekel]{tumanyan2022plugandplaydiffusionfeaturestextdriven}
Narek Tumanyan, Michal Geyer, Shai Bagon, and Tali Dekel.
\newblock Plug-and-play diffusion features for text-driven image-to-image translation.
\newblock In \emph{Proceedings of the IEEE/CVF Conference on Computer Vision and Pattern Recognition (CVPR)}, pages 1921--1930, 2023.

\bibitem[Voleti et~al.(2024)Voleti, Yao, Boss, Letts, Pankratz, Tochilkin, Laforte, Rombach, and Jampani]{voleti2024sv3d}
Vikram Voleti, Chun-Han Yao, Mark Boss, Adam Letts, David Pankratz, Dmitrii Tochilkin, Christian Laforte, Robin Rombach, and Varun Jampani.
\newblock {SV3D}: Novel multi-view synthesis and {3D} generation from a single image using latent video diffusion.
\newblock In \emph{European Conference on Computer Vision (ECCV)}, 2024.

\bibitem[Voynov et~al.(2022)Voynov, Aberman, and Cohen-Or]{voynov2022sketchguidedtexttoimagediffusionmodels}
Andrey Voynov, Kfir Aberman, and Daniel Cohen-Or.
\newblock Sketch-guided text-to-image diffusion models.
\newblock \emph{arXiv preprint arXiv:2211.13752}, 2022.

\bibitem[Wang et~al.(2022)Wang, Zhang, Zhang, Ouyang, Chen, Chen, and Wen]{wang2022pretraining}
Tengfei Wang, Ting Zhang, Bo Zhang, Hao Ouyang, Dong Chen, Qifeng Chen, and Fang Wen.
\newblock Pretraining is all you need for image-to-image translation.
\newblock \emph{arXiv:2205.12952}, 2022.

\bibitem[Wang et~al.(2004)Wang, Bovik, Sheikh, and Simoncelli]{ssimwangImageQualityAssessment2004}
Z. Wang, A.C. Bovik, H.R. Sheikh, and E.P. Simoncelli.
\newblock Image {{Quality Assessment}}: {{From Error Visibility}} to {{Structural Similarity}}.
\newblock \emph{IEEE Transactions on Image Processing}, 13\penalty0 (4):\penalty0 600--612, 2004.

\bibitem[Wu et~al.(2023{\natexlab{a}})Wu, Zhang, Fu, Wang, Jiawei~Ren, Wu, Yang, Wang, Qian, Lin, and Liu]{wu2023omniobject3d}
Tong Wu, Jiarui Zhang, Xiao Fu, Yuxin Wang, Liang~Pan Jiawei~Ren, Wayne Wu, Lei Yang, Jiaqi Wang, Chen Qian, Dahua Lin, and Ziwei Liu.
\newblock Omniobject3d: Large-vocabulary 3d object dataset for realistic perception, reconstruction and generation.
\newblock In \emph{IEEE/CVF Conference on Computer Vision and Pattern Recognition (CVPR)}, 2023{\natexlab{a}}.

\bibitem[Wu et~al.(2023{\natexlab{b}})Wu, Hao, Sun, Chen, Zhu, Zhao, and Li]{wu2023humanpreferencescorev2}
Xiaoshi Wu, Yiming Hao, Keqiang Sun, Yixiong Chen, Feng Zhu, Rui Zhao, and Hongsheng Li.
\newblock Human preference score v2: A solid benchmark for evaluating human preferences of text-to-image synthesis.
\newblock \emph{arXiv preprint arXiv:2306.09341}, 2023{\natexlab{b}}.

\bibitem[Yenphraphai et~al.(2024)Yenphraphai, Pan, Liu, Panozzo, and Xie]{yenphraphai2024imagesculptingpreciseobject}
Jiraphon Yenphraphai, Xichen Pan, Sainan Liu, Daniele Panozzo, and Saining Xie.
\newblock Image sculpting: Precise object editing with 3d geometry control.
\newblock \emph{arXiv preprint arXiv:2401.01702}, 2024.

\bibitem[Zhang et~al.(2023{\natexlab{a}})Zhang, Herrmann, Hur, Cabrera, Jampani, Sun, and Yang]{zhang2023tale}
Junyi Zhang, Charles Herrmann, Junhwa Hur, Luisa~Polania Cabrera, Varun Jampani, Deqing Sun, and Ming-Hsuan Yang.
\newblock A tale of two features: Stable diffusion complements {DINO} for zero-shot semantic correspondence.
\newblock In \emph{Thirty-seventh Conference on Neural Information Processing Systems}, 2023{\natexlab{a}}.

\bibitem[Zhang et~al.(2023{\natexlab{b}})Zhang, Rao, and Agrawala]{zhang2023addingconditionalcontroltexttoimage}
Lvmin Zhang, Anyi Rao, and Maneesh Agrawala.
\newblock Adding conditional control to text-to-image diffusion models.
\newblock In \emph{IEEE International Conference on Computer Vision (ICCV)}, 2023{\natexlab{b}}.

\bibitem[Zhang et~al.(2018)Zhang, Isola, Efros, Shechtman, and Wang]{zhang2018unreasonableeffectivenessdeepfeatures}
Richard Zhang, Phillip Isola, Alexei~A Efros, Eli Shechtman, and Oliver Wang.
\newblock The unreasonable effectiveness of deep features as a perceptual metric.
\newblock In \emph{CVPR}, 2018.

\bibitem[Zhang et~al.(2025)Zhang, Liu, Chen, and Xu]{zhang2024gooddraggoodpracticesdrag}
Zewei Zhang, Huan Liu, Jun Chen, and Xiangyu Xu.
\newblock Gooddrag: Towards good practices for drag editing with diffusion models.
\newblock In \emph{The Thirteenth International Conference on Learning Representations}, 2025.

\end{thebibliography}
}


\end{document}